\documentclass[onecolumn,11pt]{IEEEtran}
\usepackage{cite}
\usepackage{amsmath,amssymb,amsfonts}

\usepackage[colorlinks=true, linkcolor=blue, citecolor=blue, urlcolor=blue]{hyperref}

\usepackage{booktabs}
\usepackage{multirow}
\usepackage{colortbl}

\usepackage{graphicx}
\graphicspath{ {./Figures/} }

\begin{document}

\title{SEF: A Method for Computing Prediction Intervals by Shifting the Error Function in Neural Networks
\footnote{\textbf{Paper has been accepted in the \textsl{2024 International Conference on Computer and Applications} (\href{https://icca-conf.info/icca-2024}{ICCA'24}), Cairo, Egypt, December 17–19, 2024. }}}

\author{\IEEEauthorblockN{Efstratios~V.~Aretos \ and \ Dimitris~G.~Sotiropoulos}\\
        \IEEEauthorblockA{\textit{Department of Electrical and Computer Engineering} \\
                          \textit{School of Engineering, University of Peloponnese} \\ 
                          GR 263 34, Patras, Greece \\
                               Email:   \{e.aretos, dg.sotiropoulos\}@uop.gr}
        }
\maketitle

\thispagestyle{empty}

\begin{abstract}
In today's era, Neural Networks (NN) are applied in various scientific fields such as robotics, medicine, engineering, etc. However, the predictions of neural networks themselves contain a degree of uncertainty that must always be taken into account before any decision is made. This is why many researchers have focused on developing different ways to quantify the uncertainty of neural network predictions. Some of these methods are based on generating prediction intervals (PI) via neural networks for the requested target values. The SEF (Shifting the Error Function) method presented in this paper is a new method that belongs to this category of methods. The proposed approach involves training a single neural network three times, thus generating an estimate along with the corresponding upper and lower bounds for a given problem. A pivotal aspect of the method is the calculation of a parameter from the initial network's estimates, which is then integrated into the loss functions of the other two networks. This innovative process effectively produces PIs, resulting in a robust and efficient technique for uncertainty quantification. To evaluate the effectiveness of our method, a comparison in terms of successful PI generation between the SEF, PI3NN and PIVEN methods was made using two synthetic datasets. 
\end{abstract}


\section{Introduction}\label{sec:intro}

Neural networks are applied in various areas of human activity, such as the economy (e.g. prediction of stock prices or various economic indicators), medicine (e.g. prediction and diagnosis of diseases), transport (e.g. autonomous driving, safety assistance systems), etc. Neural networks have a highly successful track record in modeling complex situations and phenomena, often providing accurate answers to complex classification problems, regression, etc. However, in many cases, the predictions of neural networks involve a degree of uncertainty, which in many cases, such as in autonomous driving and medical diagnosis, can be fatal if not considered.
It is therefore no coincidence that in recent years scientific research has focused on quantifying the uncertainty of neural network predictions while ensuring that these predictions still provide satisfactory answers to given problems \cite{Abdar_2021}.

One of the most common methods for uncertainty quantification (UQ) is to use prediction intervals (PI). In practice, the neural network is used so that its final response is given in the form of intervals intended to estimate the interval within which a future observation is likely to lie. This practice is beneficial in cases where we have a regression problem in which a continuous value has to be estimated and predicted, such as an economic indicator or temperature, etc. In this way, decision making can be made more accessible, as knowing the interval range in which the final value of the phenomenon under study is expected to fluctuate makes it easier to choose the next steps or actions. Thus, using neural networks to create prediction intervals finds application in many problems of everyday life, such as sales forecasting, business risk analysis, weather forecasting, energy demand forecasting, etc.\cite{Abdar_2021}.

\subsection{Related work}
The first papers aimed at creating methods to quantify uncertainty appeared about 25 years ago \cite{Neal_1996,Hwang_1997,de_Veaux_1998}, but the number of publications has increased in recent years. Two widely used methods for uncertainty quantification are the Bayesian approximation and ensemble learning techniques \cite{Abdar_2021,Gawlikowski_2023, Khosravi_2011a}. Many articles focus on Bayesian neural networks to quantify the uncertainty associated with deep neural network predictions using variational inference, sampling approaches, or Laplace approximation \cite{Gawlikowski_2023,Jospin_2022, Wang_2016}. 
This category also includes Monte Carlo and dropout methods \cite{Neal_1996, Gal_2016}. 
Methods using ensemble techniques create forecasts based on predictions obtained from multiple members of a model ensemble \cite{Gawlikowski_2023}. 
There are empirical ensemble approaches \cite{Pearce_2020}, methods that use ensemble pruning algorithms such as deep ensemble \cite{Hu_2019}, and Bayesian ensemble learning \cite{Fersini_2014}.

More modern methods are based on the calculation of prediction intervals (PIs), which are derived from the output of a neural network, ensuring that the prediction value falls within the bounds for a specified confidence level $\gamma$. Examples of such methods include the SQR method \cite{Tagasovska_2019}, which proposes a loss function to learn all the conditional quantiles of a given target variable, and methods like LUBE \cite{Khosravi_2011b}, QD \cite{Pearce_2018}, and IPIV (or PIVEN) \cite{Simhayev_2022,Simhayev_PIVEN_2020}, which use modified loss functions containing appropriate hyperparameters to minimize the width of the lower and upper bounds while meeting the required confidence level. Furthermore, the PI3NN method \cite{Liu_PI3NN_2022} uses three neural networks to generate a point estimate, as well as the upper and lower bounds of the prediction interval.

\subsection{Motivation}
Although previous methods for finding suitable PIs, such as QD \cite{Pearce_2018}, SQR \cite{Tagasovska_2019}, and IPIV \cite{Simhayev_2022}, have been successful, they also present certain disadvantages and limitations.
\textit{i)} Often employ complex loss functions that include multiple hyperparameters to achieve optimal PIs. The extensive use of these hyperparameters typically makes these methods time-consuming, requiring precise tuning to be efficient. Determining the appropriate parameter values can be challenging, and the parameters often need to be readjusted for each problem.
\textit{ii)} Require substantial computational resources and extensive memory use. Generating appropriate PIs frequently involves complex procedures or custom architectures, increasing the computational difficulty. 
\textit{iii)} Applying these methods universally across different problems is often impractical, as each new problem may necessitate specific adjustments and redesigns of the implementation method.

All of the above has been the motivation for our research. We tried to create a new method of creating PIs without the shortcomings of the previous methods.  Our efforts led us to the SEF (Shifting the Error Function)  method.

\subsection{Objectives-Organization of the paper}
The SEF method presented in this paper focuses on using appropriate neural networks to produce satisfactory PIs for any given problem, but at the same time corrects the weaknesses of other methods such as, among others, complex architectures-structures and the existence of non-self-adjusting parameters. SEF method generates with the help of 3 neural networks suitable PIs that encapsulate the desired real targeting values satisfying a predefined confidence level $\gamma$ and have the smallest possible width in order to ensure the practical usefulness of the method. More specifically, in this paper with the presentation of the SEF method we try to present the main advantages of the SEF method which are:
\begin{itemize}
 \item its universal application to any regression problem,
 \item its ease of application as it does not require complex network architectures or complex error functions
 \item its application is not computationally expensive nor does it require complex time-consuming procedures
 \item no need to manually find the optimal value of the hyperparameter it uses, as it is calculated automatically in the process
\end{itemize}  
For this reason, the following structure is followed in this paper: in Section~\ref{sec:SEF_method} a presentation of the proposed SEF method is given first by formulating and formalizing the problem to be solved (subsection~\ref{subsec:formulation}), then by analyzing the steps of the main algorithm (\ref{SEF_algorithm}) and finally justifying the whole procedure (\ref{justify}). In Section~\ref{sec:Examples} an application of the SEF method to two synthetic datasets is conducted and the results are compared with the results of two other recent PI construction methods, the PI3NN method and the PIVEN method. In Section~\ref{sec:conclusions} we present the main conclusions we have reached in this paper and in Section~\ref{sec:future} we discuss the direction we will take in future steps.

\section{The Proposed SEF method}\label{sec:SEF_method}

\subsection{Problem Formulation}\label{subsec:formulation}
\vspace{-0.5mm}
Consider a dataset $D = \{(X_i, y_i)\}_{i=1}^n$ where each $X_i \in \mathbb{R}^d$ represents a $d$-dimensional input vector and $y_i \in \mathbb{R}$ is the corresponding target value. The objective is to construct prediction intervals (PIs) $(l_i, u_i)$ for each target value $y_i$ with a desired confidence level $\gamma \in (0, 1)$ (typically $\gamma = 0.95$ or $\gamma = 0.99$), such that the following conditions hold:

\begin{itemize}
    \item $P(l_i \le y_i \le u_i) \ge \gamma$
    \item $P(y_i < l_i) = P(y_i > u_i) = \frac{1 - \gamma}{2}$
\end{itemize}

Here, $l_i$ and $u_i \in \mathbb{R}$ represent the lower and upper bounds of the PI for each $y_i$, respectively. The desired confidence level $\gamma$ indicates the probability that the true value $y_i$ falls within the interval $(l_i, u_i)$.
Define a vector $K = \{k_i\}$, where $k_i$ is an indicator function defined \cite{Pearce_2018} as
\[
k_i = \begin{cases} 
1, & \text{if } l_i \le y_i \le u_i \\
0, & \text{otherwise}.
\end{cases}
\] 
The vector $K$ essentially denotes whether each target value $y_i$ is encapsulated within its corresponding PI.
The \textit{prediction interval coverage probability} (PICP) \cite{Pearce_2018} quantifies the proportion of target values $(y_i)$ prediction contained within their respective PIs. It is calculated as follows:
\[
\text{PICP} = \frac{1}{n} \sum_{i=1}^n k_i
\]
In practice, our objective is to ensure that $PICP \ge \gamma$. Furthermore, the \textit{mean prediction interval width} (MPIW)\cite{Pearce_2018} is defined as
\[
\text{MPIW} = \frac{1}{n} \sum_{i=1}^n (u_i - l_i)
\]
which measures the average width of the PIs and should be minimized, provided that the condition $PICP \ge \gamma$ is satisfied. Another measure that is used is the Normalized MPIW (NMPIW),
\[
\text{NMPIW} = \frac{MPIW}{R} 
\]
where $R$ corresponds to the range of target values and allows us to compare PIs from different datasets \cite{Khosravi_2011a}. The goal is to construct prediction intervals $(l_i, u_i)$ that achieve a high coverage probability ($\ge \gamma$) while keeping the intervals as narrow as possible.

\subsection{The SEF Algorithm}\label{SEF_algorithm}

The proposed method involves four main steps that require the training of three neural networks (NN). These networks share the same architecture, differing only in the constants added to their loss functions to achieve different objectives. The primary advantage of this approach is that it simplifies the implementation and enhances execution speed by avoiding extensive hyperparameter tuning beyond the initial network's essential parameters. In what follows, we describe the four steps of the SEF algorithm. 

\bigskip\noindent
\textit{Step 1: Training the initial estimator}. 
This step involves training a neural network to solve a regression problem by approximating the target values $(y_i)$.
Let ${\text{NN}}_{\text{approx}}$ be a neural network with an architecture defined by the number of layers, nodes within each layer, activation functions, and other hyperparameters. This network produces predictions $\widehat{y}_i = {\text{NN}}_{\text{approx}}(X_i; \theta)$, while the network parameters $\theta$ are optimized by minimizing the mean squared error (MSE) loss function
$$
L(\theta) = \frac{1}{n} \sum_{i=1}^n \left( {\text{NN}}_{\text{approx}}(X_i; \theta) - y_i \right)^2.
$$
The goal is to train the first neural network to accurately approximate the target values $\widehat{y}$, where $\widehat{y} = {\text{NN}}_{\text{approx}}(X)$. To achieve this, it is essential to prevent overfitting and ensure generalization. Various regularization techniques, such as dropout and weight decay, along with validation methods, are employed. In this paper, a validation set $D_{\text{valid}} \subset D$ is used to monitor the performance of the network during training and to fine-tune the model.

\bigskip\noindent
\textit{Step 2: Determination of the displacement constant $\mu$}.  
Compute the residuals $d_i = \widehat{y}_i - y_i$ for each observation $1 \leq i \leq n$. These residuals represent the differences between the predicted values $\widehat{y}_i$ and the actual target values $y_i$. Sort the residuals in ascending order to obtain the sequence $\delta_1, \delta_2, \ldots, \delta_n$. Define the indices $m_1$ and $m_2$ corresponding to the quantiles $(1-\gamma)/2$ and $(1+\gamma)/2$, respectively, as follows:
$$
m_1 = \left\lfloor \frac{(1 - \gamma)}{2} \cdot n \right\rfloor \quad \text{and} \quad m_2 = \left\lceil \frac{(1 + \gamma)}{2} \cdot n \right\rceil.
$$
These indices correspond to the quantiles $(1-\gamma)/2$ and $(1+\gamma)/2$, respectively, ensuring that at least \(\gamma\) proportion of the residuals lie within the central region, while the extremes are flagged as potential outliers.
The floor ($\left\lfloor \cdot \right\rfloor$) and ceiling ($\left\lceil \cdot \right\rceil$) functions ensure that the indices $m_1$ and $m_2$ are valid integers within the range of the dataset. In cases where after this procedure we conclude that  $m_{1} = 0$ we designate $m_{1} = 1$. 
The displacement constant $\mu$ is then defined as:
$$
\mu = \max \left( |\delta_{m_1}|, |\delta_{m_2}| \right).
$$
Absolute values ensure that $\mu$ accurately reflects the maximum deviation, regardless of whether the residuals are positive or negative. The absolute value of these differences indicates the distance between the actual value $y_i$ and the predicted value $\widehat{y}_i$. Specifically, if the actual value $y_i$ is higher than the predicted value $\widehat{y}_i$, the residual $d_i$ will be negative ($d_i < 0$), and if it is lower, the residual $d_i$ will be positive ($d_i > 0$). We use absolute values to ensure that both underestimations and overestimations are taken into account when determining the displacement constant $\mu$.

\bigskip\noindent
\textit{Step 3: Training the lower bound estimator}.
In this step, ${\text{NN}}_{\text{lower}}$ is trained to approximate the lower bound $l_i$ of the PI, i.e., $l_i = {\text{NN}}_{\text{lower}}(X)$. The architecture and parameter values (number of layers, nodes, activation function) are the same as ${\text{NN}}_{\text{approx}}$ from Step 1, but it uses a modified MSE loss function:
$$
L_{\text{lower}}(\theta) = \frac{1}{n} \sum_{i=1}^n \left[ {\text{NN}}_{\text{lower}}(X_i; \theta) - \left( y_i - \mu\right) \right]^2.
$$
This training reduces the error according to the constraint provided by $\mu$. To avoid overfitting, a validation set $D_{\text{valid}} \subset D$ is used, ensuring that the network is well generalized to unseen data.

\bigskip\noindent
\textit{Step 4: Training the upper bound estimator}.
This step involves training ${\text{NN}}_{\text{upper}}$ to approximate the upper bound $u_i$ of the PI. Like ${\text{NN}}_{\text{lower}}$, this network has the same architecture and parameter values as ${\text{NN}}_{\text{approx}}$, but uses another modified MSE loss function:
$$
L_{\text{upper}}(\theta) = \frac{1}{n} \sum_{i=1}^n \left[ {\text{NN}}_{\text{upper}}(X_i; \theta) - \left(y_i + \mu\right) \right]^2.
$$
The network ${\text{NN}}_{\text{upper}}$ is trained to minimize this loss, ensuring that the upper bound $u_i$ appropriately encapsulates the target values and a validation set  $D_{\text{valid}}$  is utilized to prevent overfitting.

\subsection{Justification}\label{justify}

To understand the justification for our algorithm from a geometrical perspective, let us depict pairs $(y_i, \widehat{y_i})$ in a coordinate system, where the horizontal axis represents the actual target values $y_i$, and the vertical axis represents the predicted values $\widehat{y_i}$. By analyzing this illustration, we can grasp the concept behind the SEF algorithm.

The bisector of the first and third quadrants is the line where $y = \widehat{y_i}$. This line represents perfect predictions, where the predicted value exactly matches the target value. Any point on this line indicates that the prediction error is zero.

Points above the line $y = \widehat{y_i}$ correspond to instances where the predicted value $\widehat{y_i}$ is greater than the actual target value $y_i$ ($\widehat{y_i} > y_i$). This indicates that the model is overestimating the target value. In contrast, points below the line $y = \widehat{y_i}$ correspond to instances where the predicted value $\widehat{y_i}$ is less than the actual target value $y_i$ ($\widehat{y_i} < y_i$). This indicates that the model is underestimating the target value.

The absolute difference $|d_i| = |\widehat{y_i} - y_i|$ determines the vertical distance of a point $(y_i, \widehat{y_i})$ from the bisector $y = \widehat{y_i}$. This distance quantifies the prediction error. A smaller distance indicates a more accurate prediction, whereas a larger distance indicates a larger prediction error.

\begin{figure}[htb!]
  \centering
  \includegraphics[width=0.8\linewidth]{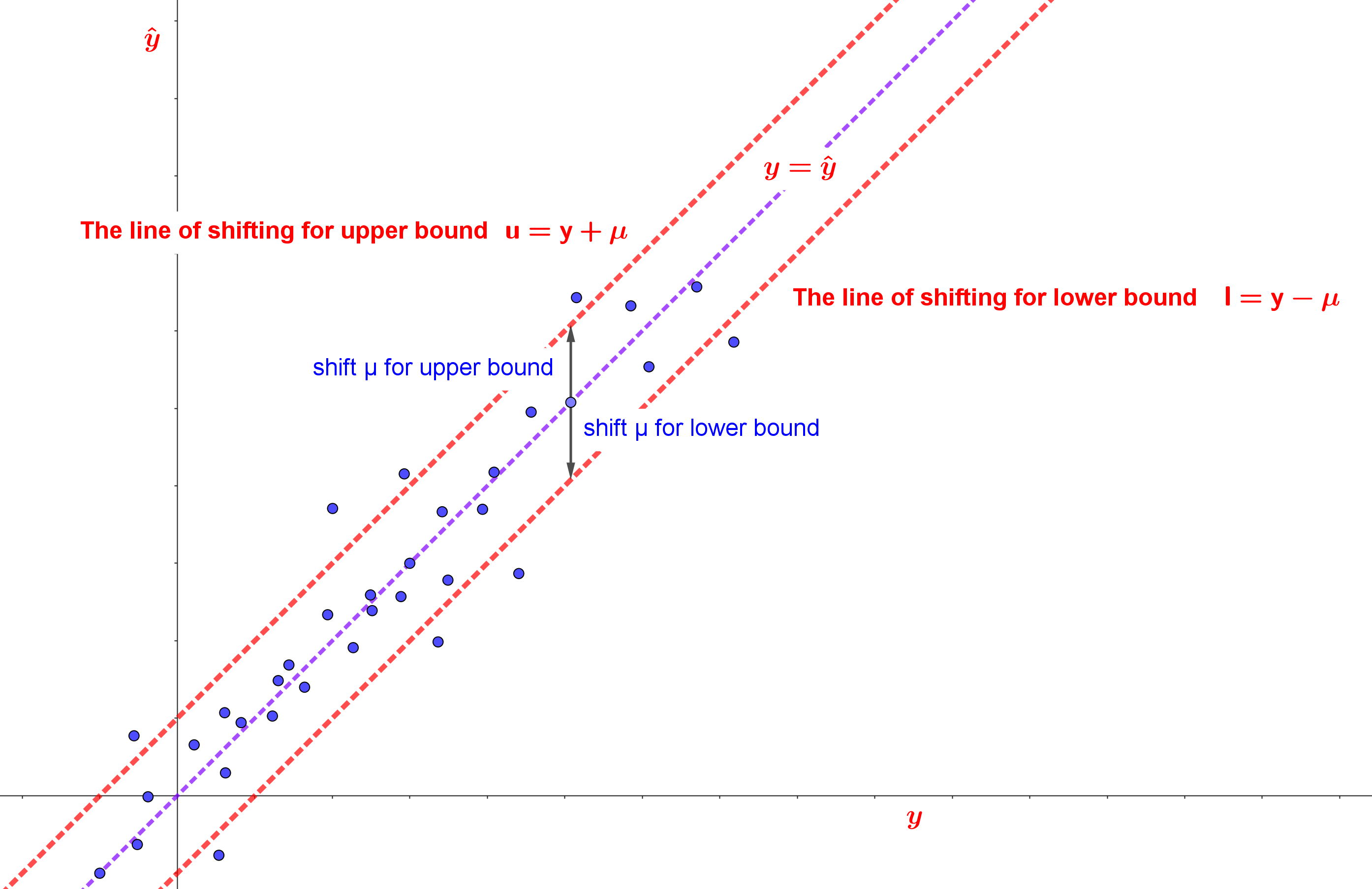}
  \caption{Geometrical representation of the lines of shifting for the upper and lower bounds.}
  \label{fig1:ShiftingIdea}
\end{figure}

Figure~\ref{fig1:ShiftingIdea} illustrates the geometrical concept of shifting lines for the upper and lower bounds. The bisector line $y = \widehat{y_i}$ represents where the predicted values exactly match the actual values. The lines $u = y + \mu$ and $l = y - \mu$ represent the upper and lower bounds of the prediction intervals, respectively. These lines are shifted by $\mu$ units from the bisector, ensuring that the prediction intervals encapsulate the actual target values with a specified confidence level $\gamma$. This visual representation helps us to understand how the SEF algorithm adjusts the bounds to achieve accurate and reliable prediction intervals.

During the training process, the differences $d_i = \widehat{y_i} - y_i$ should become progressively smaller. This implies that the points $(y_i, \widehat{y_i})$ in the plane should move closer and closer to the bisector $y = \widehat{y_i}$. At the end of the network training, these points should ideally be located at the shortest possible distance from the bisector, indicating minimal prediction errors.

However, to generate the PIs, we need to find two values, $l_i$ and $u_i$, for each $y_i$ such that $l_i \leq y_i \leq u_i$. This translates to $l_i - y_i \leq 0$ and $u_i - y_i \geq 0$. In the aforementioned coordinate plane, the points $(y_i, l_i)$ should be below the bisector, while the points $(y_i, u_i)$ should be above the bisector $y = \widehat{y_i}$. We can formalize this by introducing $\varepsilon_i, \varepsilon_i' \in \mathbb{R}$ such that for each $1 \leq i \leq n$, the following hold:
\begin{equation}\label{eq:lines}
l_i = y_i - \varepsilon_i  \quad \text{and} \quad u_i = y_i + \varepsilon_i'.
\end{equation}
To generate PIs, we choose a suitable constant $\mu$ such that $P[y_i - \mu \leq y_i \leq y_i + \mu] \geq \gamma$ for each $1 \leq i \leq n$. Setting $\varepsilon_i = \varepsilon_i' = \mu$ for each $1 \leq i \leq n$, Eqs~(\ref{eq:lines}) become $l_i = y_i - \mu$ and $u_i = y_i + \mu$. These equations represent two lines in the plane, resulting from the vertical displacement of the bisector $y = \widehat{y_i}$ by $\mu$ units downward and upward, respectively (see Figure~\ref{fig1:ShiftingIdea}).

Thus, the problem of creating appropriate PIs reduces to finding the constant $\mu$. The proposed PI-SEF method uses the trained network ${\text{NN}}_{\text{approx}}$ to determine $\mu$ using residuals $d_i = \widehat{y_i} - y_i$. The value of $\mu$ is chosen so that $\gamma \%$ of the $y_i$ values fall within the interval $[y_i - \mu, y_i + \mu]$. This requirement can be mathematically expressed using the indicator function $k_i$ as follows:
$$
\frac{1}{n} \sum_{i=1}^{n} k_i = \gamma,
$$
where $k_i = 1$ if $l_i \le y_i \le u_i$ and $k_i = 0$ otherwise. This ensures that the proportion of target values $y_i$ encapsulated within their respective prediction intervals is at least $\gamma$.

Training of the neural networks ${\text{NN}}_{\text{lower}}$ and ${\text{NN}}_{\text{upper}}$ aims to obtain the lower and upper bounds $l_i$ and $u_i$ of the PI, respectively. The loss functions used in these networks ensure that $y_i$ stays within the corresponding PIs with the chosen confidence level $\gamma \%$ and that the PI widths are minimized.

\begin{figure}[htbp]
  \centering
  \includegraphics[width=0.95\linewidth]{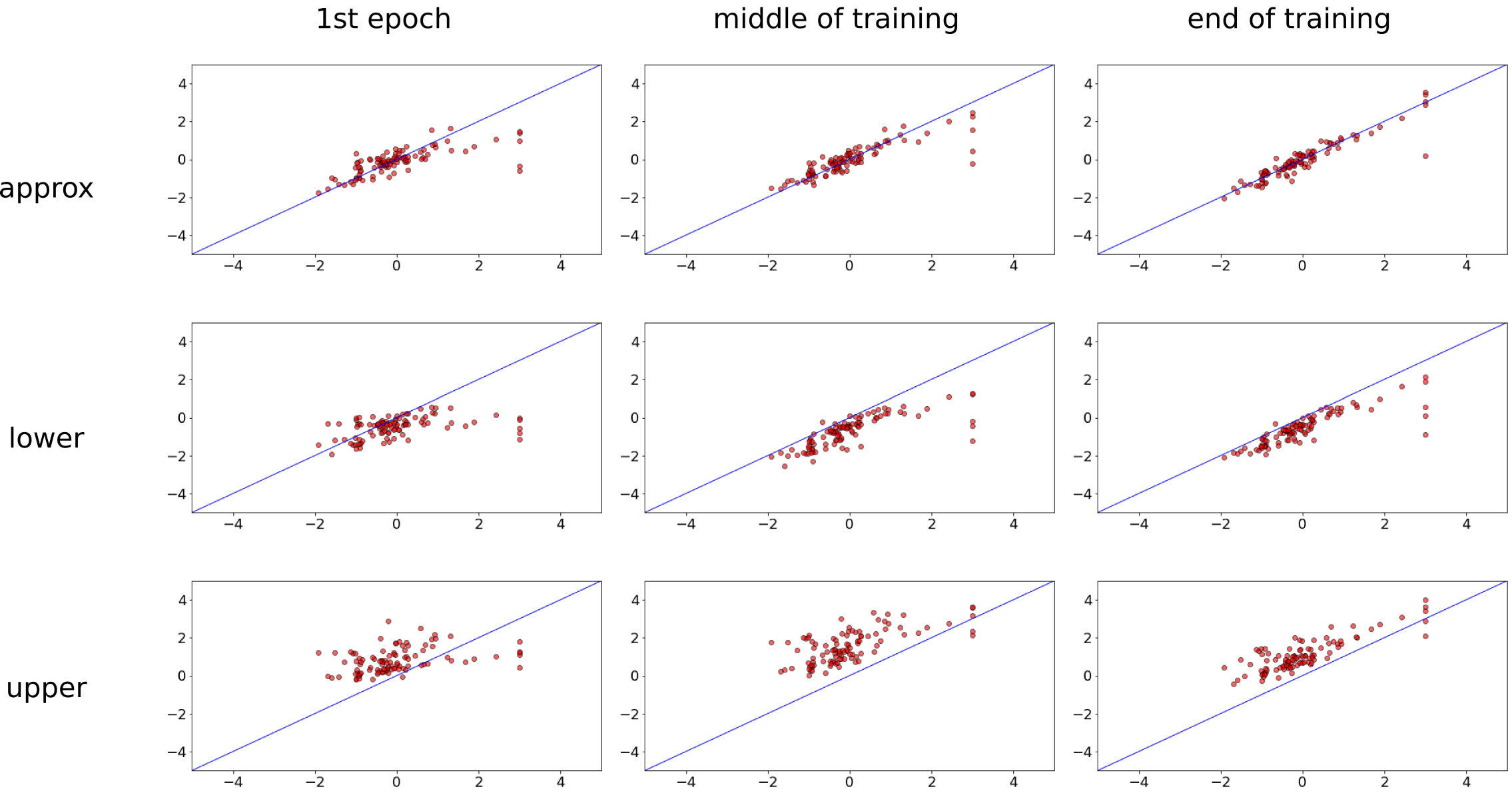}
  \caption{Graphical representations of the alignment of points at the beginning, during, and at the end of the training process.}
  \label{fig:training_epochs}
\end{figure}

Geometrically, during training, the points $(y_i, u_i)$ should gradually move to align with the line $u = y + \mu$, placing these points in the half-plane where $y < \widehat{y_i}$. Similarly, the points $(y_i, l_i)$ should move to align with the line $l = y - \mu$, placing them in the half-plane where $y > \widehat{y_i}$.

Figure~\ref{fig:training_epochs} illustrates this change during the training process. The graphical representations show the alignment of the points at the beginning, during, and at the end of the training. As training progresses, the points $(y_i, u_i)$ and $(y_i, l_i)$ move closer to the desired lines $u = y + \mu$ and $l = y - \mu$, respectively, indicating that the prediction intervals are being properly adjusted to encapsulate the actual target values with the specified confidence level $\gamma$.

\section{Examples}\label{sec:Examples}
In this section, we present the comparative results of the application of the SEF method against two other recent and well-established methods, the PIVEN method and the PI3NN method. Our aim is to use two synthetic datasets (homoscedastic and heteroscedastic) to demonstrate the reliability and effectiveness of the SEF method by comparing it with the two other methods. We report PICP, MPIW, and NMPIW as performance metrics. PICP measures the percentage of true values that fall within the prediction intervals, indicating their reliability, while MPIW and NMPIW measure the average width of the prediction intervals, reflecting their accuracy.
For all examples and runs, we set the confidence level to $\gamma = 0.95$.

\subsection{Trigonometric Function with Varying Noise Levels}
In this example, we evaluated the effectiveness of the SEF method on the function 
$f(x) = 1.5 \cdot \sin(x) + \varepsilon,$ where $\varepsilon$ is a normally distributed noise component with varying standard deviations. We created five datasets with varying levels of noise, denoted by $\epsilon \sim \mathcal{N}(0, \sigma^{2})$, where $\sigma^{2}$ ranged from 0.1 to 0.5 in increments of 0.1. Each of the 5 datasets comprised 1000 observations from the function $f(x)$ where $x \in [-2\pi, 2\pi]$. We first applied the SEF method with 5-fold cross-validation for each of the 5 initially created datasets. Then, we compared the SEF, PI3NN, and PIVEN methods by applying them to each of the 5 initially created datasets using a random split in train and test set.
\subsubsection{Standalone apply of the SEF method }

To evaluate the SEF method, we employed 5-fold cross-validation. This method divides the dataset into five equal-sized folds, using four folds for training and the remaining fold for testing. This process was repeated five times, ensuring each fold was the testing set exactly once. Performance metrics were averaged over these repetitions to obtain the final results. This approach mitigates the effects of data variability and provides a more precise estimate of the model's performance. 

The neural networks used had two hidden layers with 100 and 50 nodes, respectively, equipped with the ReLU activation function. The optimization algorithm ``Adam'' was used, and 10\% of the training set was used as a validation set for early stopping. We used the following metrics to evaluate the SEF method: a) PICP, b) MPIW, and c) $2\mu$, which is twice the shifting constant $\mu$ used in the SEF method and serves as a basis for comparison with MPIW.
\begin{table}[htbp]
\caption{Performance metrics of SEF method using 5-Fold Cross-Validation on $f(x) = 1.5  \sin(x) + \varepsilon$ with varying Gaussian noise levels}
\label{tab:sef-5fold-cv}
\centering
\begin{tabular}{c|c|cccc}
\toprule
Noise  & Fold & Shift Const. $\mu$ & PICP & MPIW & $2\mu$ \\
\midrule
\multirow{6}{*}{0.1} 
    & 1 & 0.235 & 0.97 & 0.440 & 0.4696 \\
    & 2 & 0.209 & 0.95 & 0.440 & 0.4173 \\
    & 3 & 0.225 & 0.99 & 0.500 & 0.450 \\
    & 4 & 0.199 & 0.96 & 0.453 & 0.397 \\
    & 5 & 0.247 & 0.95 & 0.427 & 0.493 \\
    & \textbf{Mean} & \textbf{0.223} & \textbf{0.964} & \textbf{0.452} & \textbf{0.446} \\
\midrule
\multirow{6}{*}{0.2} 
    & 1 & 0.395 & 0.955 & 0.810 & 0.790 \\
    & 2 & 0.401 & 0.965 & 0.845 & 0.802 \\
    & 3 & 0.451 & 0.960 & 0.851 & 0.901 \\
    & 4 & 0.402 & 0.940 & 0.791 & 0.803 \\
    & 5 & 0.384 & 0.960 & 0.743 & 0.768 \\
    & \textbf{Mean} & \textbf{0.407} & \textbf{0.956} & \textbf{0.808} & \textbf{0.813} \\
\midrule
\multirow{6}{*}{0.3} 
    & 1 & 0.626 & 0.935 & 1.240 & 1.252 \\
    & 2 & 0.571 & 0.945 & 1.143 & 1.142 \\
    & 3 & 0.634 & 0.970 & 1.353 & 1.267 \\
    & 4 & 0.705 & 0.965 & 1.235 & 1.410 \\
    & 5 & 0.669 & 0.985 & 1.369 & 1.337 \\
    & \textbf{Mean} & \textbf{0.641} & \textbf{0.960} & \textbf{1.268} & \textbf{1.281} \\
\midrule
\multirow{6}{*}{0.4} 
    & 1 & 0.830 & 0.935 & 1.599 & 1.660 \\
    & 2 & 0.921 & 0.980 & 1.846 & 1.842 \\
    & 3 & 0.876 & 0.955 & 1.703 & 1.751 \\
    & 4 & 0.874 & 0.965 & 1.826 & 1.747 \\
    & 5 & 0.843 & 0.965 & 1.710 & 1.686 \\
    & \textbf{Mean} & \textbf{0.869} & \textbf{0.960} & \textbf{1.737} & \textbf{1.737} \\
\midrule
\multirow{6}{*}{0.5} 
    & 1 & 1.028 & 0.945 & 2.084 & 2.055 \\
    & 2 & 1.002 & 0.945 & 1.986 & 2.003 \\
    & 3 & 1.071 & 0.955 & 2.044 & 2.141 \\
    & 4 & 1.091 & 0.975 & 2.152 & 2.183 \\
    & 5 & 1.022 & 0.935 & 1.993 & 2.043 \\
    & \textbf{Mean} & \textbf{1.043} & \textbf{0.951} & \textbf{2.052} & \textbf{2.085} \\
\bottomrule
\end{tabular}%
\end{table}
The results presented in Table~\ref{tab:sef-5fold-cv} demonstrate the robustness of the SEF method at different noise levels. The SEF method consistently achieved high PICP values across all datasets, with mean PICP values ranging from $0.951$ to $0.964$.  This indicates that the method reliably captures the true values within the prediction intervals, even as the noise level increases.As the noise level increased, the MPIW also increased as expected, reflecting the increasing uncertainty in the data. For example, the mean MPIW values were $0.452$, $0.808$, $1.268$, $1.737$, and $2.052$ for datasets with noise standard deviations of $0.1$, $0.2$, $0.3$, $0.4$, and $0.5$, respectively. 
These results show that the SEF technique efficiently adjusts the prediction intervals to retain good coverage probability across a range of noise levels, although at the expense of increased interval width.

The metric $2\mu$ provides a comparable measure to MPIW, showing similar trends and supporting the robustness of the SEF method. Furthermore, we observe that the MPIW is directly proportional to $\mu$ and is constant across all prediction intervals given the symmetry around the point estimates $\widehat{y}_i$. Especially at higher noise levels, the mean $2\mu$ value is slightly higher than the mean MPIW. 

\subsubsection{Comparing SEF, PI3NN and PIVEN methods}
To evaluate the performance of the SEF method, we also compared it with two other recent PIs generation methods, the PI3NN method and the PIVEN method. More specifically, we divided each of the five datasets with varying levels of noise of this example, randomly into Train and Test set at a rate of 80\%-20\% and applied for each of these pairs the three methods as mentioned above. In all 3 methods, the neural networks used had 2 hidden layers of 100 and 50 nodes, respectively, and the ReLU function was the activation function. The optimization algorithm ``Adam'' was used and 10\% of the training set was used as a validation set for early stopping. Especially for the PIVEN method, which requires the use of manually adjustable hyper-parameters, the following values of hyper-parameters were given $\lambda=15$, $s=160$, $\alpha=0.05$ and $\beta=0.5$. We used the PICP and NMPIW as metrics to evaluate the methods and the results are presented in Table~\ref{tab:comparative-results}.
\begin{table}[htbp]
\caption{Comparative Results of SEF, PI3NN and PIVEN methods}
\label{tab:comparative-results}
\centering
\begin{tabular}{c|cc|cc|cc}
\toprule
\multirow{2}{*}{\begin{tabular}[c]{@{}c@{}}Noise\\ $\sigma^2$\end{tabular}} & \multicolumn{2}{c|}{SEF} & \multicolumn{2}{c|}{PI3NN} & \multicolumn{2}{c}{PIVEN} \\
 & PICP & NMPIW & PICP & NMPIW & PICP & NMPIW \\
\midrule
0.1 & \textbf{0.975} & \textbf{0.131} & 0.965 & 0.179 & 0.960 & 0.345 \\
0.2 & \textbf{0.965} & \textbf{0.220} & 0.950 & 0.870 & 0.955 & 0.364 \\
0.3 & \textbf{0.965} & \textbf{0.290} & 0.915 & 0.858 & 0.955 & 0.389 \\
0.4 & 0.960 & \textbf{0.384} & \textbf{0.980} & 0.523 & 0.955 & 0.459 \\
0.5 & \textbf{0.955} & \textbf{0.394} & 0.950 & 0.778 & 0.940 & 0.445 \\
\midrule
Mean & \textbf{0.964} & \textbf{0.284} & 0.952 & 0.642 & 0.953 & 0.401 \\
\bottomrule
\end{tabular}%
\end{table}
The SEF method outperforms the other two methods in 4 of the five datasets in terms of the largest PICP value and in all datasets in terms of the smallest average value of PIs width. 

We used Friedman's rank test, a nonparametric statistical test, to assess the statistical significance of the differences between the 3 methods in terms of both PICP and NMPIW across the five datasets with varying additive Gaussian noise levels. For the PICP metric, Friedman's test yielded a test statistic of 5.2 with a corresponding $p$-value = 0.0743, which exceeds the predetermined significance level of 5\%, so we cannot conclude that there are statistically significant differences among the three methods in terms of PICP.

In contrast, for the NMPIW metric, Friedman's test yielded a test statistic of $8.4$ and $p$-value $= 0.015 < 0.05$, so we can reject the null hypothesis, suggesting statistically significant differences among the methods in terms of NMPIW. Therefore, we proceeded with post-hoc analyses to identify specific pairwise differences. 
Nemenyi test yielded $p$-values of 0.073 for SEF vs. PIVEN, 0.523 for SEF vs. PI3NN, and 0.527 for PIVEN vs. PI3NN, so none of these pairwise comparisons reached statistical significance at the 5\% level. In the sequel, we conducted Dunn's test as an additional post-hoc analysis to further investigate potential pairwise differences. Interestingly, this test revealed a significant difference between SEF and PI3NN ($p$-value = 0.024), while the comparisons of SEF vs. PIVEN ($p$-value = 0.258) and PI3NN vs. PIVEN ($p$-value = 0.258) remained non-significant.

\subsection{Time-Dependent Sinusoidal Noise}
In this heteroscedastic example, we evaluated the effectiveness of the SEF method using a function with variable noise. Specifically, we used the function
$g(t) = 10 + 3t + t \cdot \sin(2t) + \epsilon_t,$ where $\epsilon_t$ is a noise component normally distributed with $\epsilon_t \sim \mathcal{N}(0, t + 0.01)$ \cite{MathWorks}. We generate 500 observations for the function $g(t)$, where $t \in [0, 4\pi]$. The dataset was randomly divided into a training set and a test set in a ratio 80\% to 20\%. In our first experiment, we trained the neural networks using the SEF method on the training set and evaluated it on the test set. This process was repeated five times with different random splits, and the results were recorded. In our second experiment, we followed the same procedure of randomly separating the dataset into a training set and test set, but this time, we applied all 3 methods of the previous example (SEF, PI3NN, and PIVEN) to compare their results. We continued in a third experiment and used a 5-fold cross-validation on the original dataset for the SEF method alone, and in the fourth experiment, a 10-fold cross-validation on the original dataset for each of the 3 methods in order to further evaluate the SEF method and its effectiveness in comparison to the other methods.  Performance metrics were averaged over the five and ten folds to obtain the final results.

For all the experiments that followed and for all of the methods, the neural networks that were used had two hidden layers with 200 and 100 nodes, respectively, equipped with the ReLU activation function. The optimization algorithm ``Adam'' was used, and 10\% of the training set was used as a validation set. The following values of hyper-parameters were given for the PIVEN method:  $\lambda=15$, $s=160$, $\alpha=0.05$, and $\beta=0.5$.

\subsubsection{Results of SEF method on Random Dataset Splits} 
The original dataset was divided into five random pairs of training and test sets in an 80\% to 20\% ratio. The SEF method was applied and evaluated for each of these pairs. 
\begin{table}[htbp]
    \caption{Results for 5 Random Splits of the Dataset}
    \label{tab:example2_dataset1}
    \centering
    \begin{tabular}{ccccc}
        \toprule
        Split & Shift Const. $\mu$ & PICP & MPIW & $2\mu$ \\
        \midrule              
        1 & 34.093 & 0.980 & 12.872 & 68.186  \\
        2 & 33.859 & 0.970 & 12.673 & 67.718  \\
        3 & 33.733 & 0.980 & 12.191 & 67.466  \\
        4 & 34.117 & 0.990 & 13.219 & 68.234  \\
        5 & 33.924 & 0.990 & 12.607 & 67.848  \\
        \midrule
        Mean & 33.945 & 0.982 & 12.712 & 67.890  \\
        \bottomrule
    \end{tabular}
\end{table}
The results in Table~\ref{tab:example2_dataset1} show a high PICP with a mean of $0.982$, confirming that the SEF method is effective in capturing true values within the prediction intervals, achieving the required confidence level of $\gamma = 0.95$. The MPIW is $12.71$, ranging from $12.19$ to $13.22$. 

\begin{figure}[htbp]
  \centering
  \includegraphics[width=0.9\linewidth]{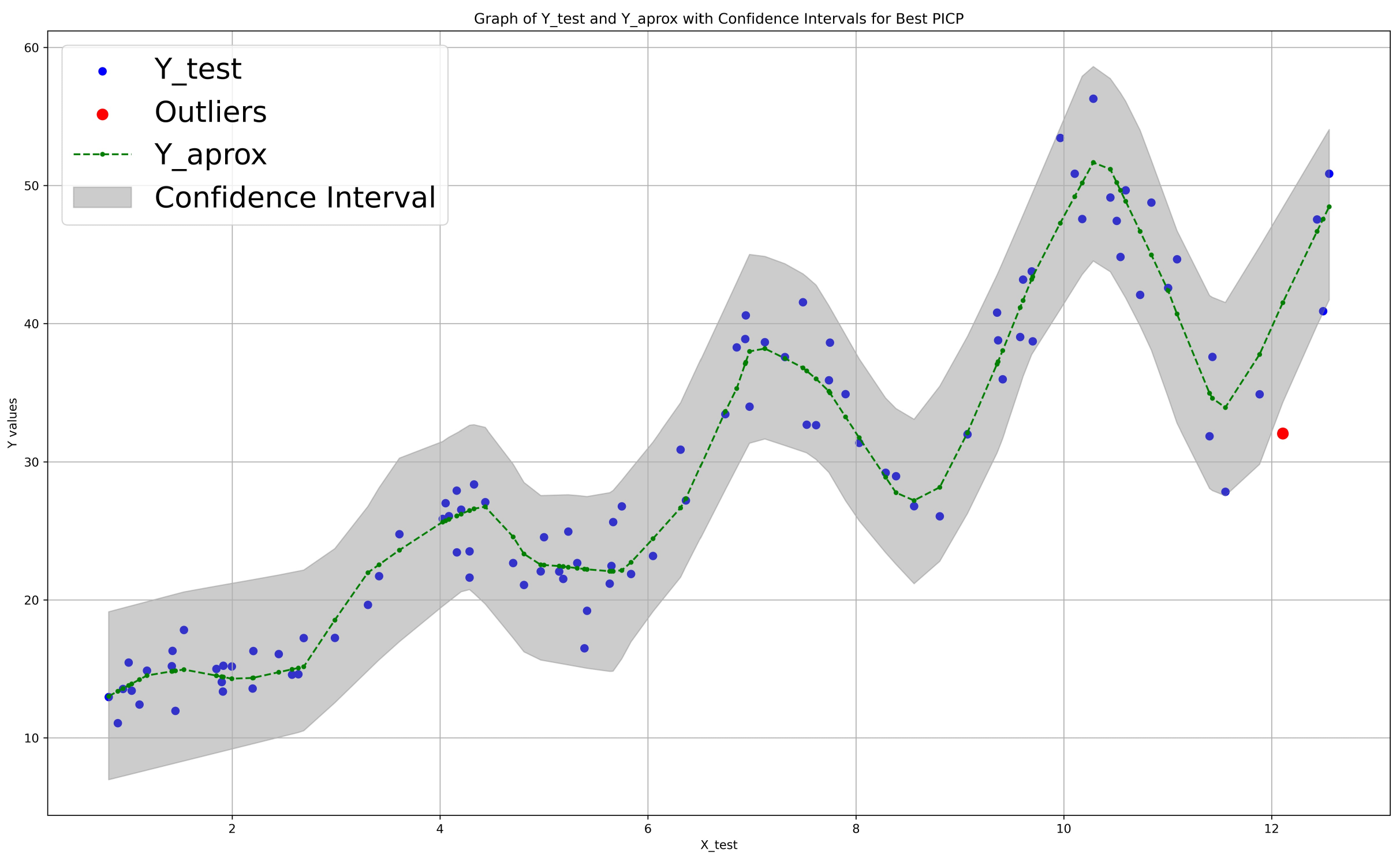}
  \caption{Scatter plot illustrating the highest PICP case from the fourth random split.}
  \label{fig:bestPICP}
\end{figure}

Figure~\ref{fig:bestPICP} illustrates the actual target values ($y_i$) versus the predicted values ($\widehat{y_i}$), along with the lower and upper bounds of the prediction intervals (PIs) for the fourth split. The grey zone in the graph represents these prediction intervals, indicating the range within which true values are expected to fall with a specified confidence level. The outlier, highlighted in red, demonstrates the robustness of the SEF method in maintaining a high coverage probability even in the presence of anomalies in the data.

\subsubsection{Comparison on Random Dataset Splits} 
In this experiment we followed exactly the same procedure of the previous experiment but we applied to the created dataset splits not only the SEF method but also the PIVEN and PI3NN methods. So we randomly split the original created dataset into train and test set with a percentage of 80\% and 20\% and then we applied all 3 methods and recorded the results, namely PICP and the NMPIW.  The procedure was repeated 5 times and the results are presented in Table~\ref{tab:comparative-results-sinusoidal}.

\begin{table}[htbp]
\caption{Comparative results of SEF, PI3NN and PIVEN methods for time-dependent sinusoidal noise function}
\label{tab:comparative-results-sinusoidal}
\centering
\begin{tabular}{c|cc|cc|cc}
\toprule
\multirow{2}{*}{\begin{tabular}[c]{@{}c@{}}Random\\Split\end{tabular}} & \multicolumn{2}{c|}{SEF} & \multicolumn{2}{c|}{PI3NN} & \multicolumn{2}{c}{PIVEN} \\
 & PICP & NMPIW & PICP & NMPIW & PICP & NMPIW \\
\midrule
1 & \textbf{0.960} & \textbf{0.415} & 0.950 & 0.463 & 0.930 & 0.365 \\
2 & \textbf{0.970} & \textbf{0.289} & 0.930 & 0.438 & 0.950 & 0.407 \\
3 & 0.960 & \textbf{0.354} & \textbf{0.990} & 0.525 & 0.980 & 0.432 \\
4 & \textbf{0.980} & 0.365 & 0.960 & 0.456 & 0.960 & \textbf{0.360} \\
5 & \textbf{1.000} & 0.402 & 0.940 & \textbf{0.397} & 0.950 & 0.395 \\
\midrule
Mean & \textbf{0.974} & \textbf{0.365} & 0.954 & 0.456 & 0.954 & 0.392 \\
\bottomrule
\end{tabular}
\end{table}

As we can observe in Table~\ref{tab:comparative-results-sinusoidal}, the SEF method outperforms the PI3NN and PIVEN methods in 4 of the 5 datasets by exhibiting both the highest
PICP and the smallest width of the constructed PIs.
The mean values across the 5 runs further show the superiority of the SEF method as it not only outperforms the PICP with an average of 0.974 against 0.954 of the other two methods but also shows the lowest NMPIW value of 0.365 against 0.456 for the PI3NN method and 0.392 for the PIVEN method. 

This comparison reveals an interesting finding: whereas it is commonly expected that greater PICP values correlate with increasing NMPIW, the SEF method resists this trend in several instances, managing to have not only the best PICP but also the lowest or close to the lowest NMPIW at the same time. 
As shown in random splits 1 and 2, the SEF shows the best PICP and, at the same time, the lowest NMPIW, while in random split 4,
SEF yields a PICP of 0.98 with an NMPIW of 0.365, whereas the next best performer, PIVEN, shows a lower PICP of 0.96 with only a marginal reduction in NMPIW (0.360). To further assess the statistical significance of these observed differences, we employed Friedman's ranking method. The analysis yielded $p$-values of 0.2415 and 0.0743 for PICP and NMPIW, respectively, and therefore, there is no statistically significant difference among the three methods when considering both PICP and NMPIW across all datasets.

\subsubsection{K-Fold Cross-Validation Results for SEF Method}
We further evaluated the SEF method using a 5-fold cross-validation approach. The dataset was randomly shuffled and partitioned into five subsets. Each subset was used as the test set once, while the remaining subsets were used for training. The results of the 5-fold cross-validation are presented in Table~\ref{tab:5-fold cross-validation method}.
\begin{table}[htbp]
    \caption{Results of the SEF method using the 5-fold cross-validation}
    \label{tab:5-fold cross-validation method}
    \centering
    \begin{tabular}{ccccc}
        \toprule
        Fold & Shift Const. $\mu$  & PICP & MPIW  & \textbf{$2\mu$} \\
        \midrule
        1 & 33.569 & 0.960 & 10.749 & 67.137 \\
        2 & 34.351 & 0.990 & 12.387 & 68.702 \\
        3 & 33.759 & 0.970 & 11.816 & 67.517 \\
        4 & 33.752 & 0.940 & 11.658 & 67.504 \\
        5 & 34.482 & 0.960 & 13.168 & 68.964 \\
        \midrule
        Mean & 33.983 & 0.964 & 11.956 & 67.965 \\
        \bottomrule
    \end{tabular}
\end{table}
The mean PICP for 5-fold cross-validation is $0.964$, and the mean MPIW is $11.96$. 
The SEF method managed to achieve high PICP values, significantly higher than the target $\gamma = 0.95$ in all except one case at the 4th fold where it was slightly below $0.95$. 
This indicates that the SEF method shows robust performance in generating reliable prediction intervals even with variable noise levels. A notable observation from the results is that the values of $2\mu$ are consistently much higher than the MPIW, almost 6 times. This can be explained by the high levels of noise in the data. The SEF method adjusts the prediction intervals to maintain a high coverage probability (PICP), and the displacement constant $\mu$ is used to widen the intervals sufficiently to encapsulate the true values, ensuring that the prediction intervals are wide enough to achieve the desired coverage even in the presence of significant noise. 

\subsubsection{K-Fold Cross-Validation Comparison}
In this last experiment we use a 10-fold cross validation comparison of the methods SEF, PI3NN and PIVEN. As before the original dataset was randomly shuffled and
partitioned into ten subsets. Each subset was used as the test set once, while the remaining subsets were used for training. The results of the 10-fold cross-validation are presented in Table~\ref{tab:10-fold}.
\begin{table}[htbp]
\caption{Comparative results of SEF, PI3NN and PIVEN methods}
\label{tab:10-fold}
\centering
\begin{tabular}{c|cc|cc|cc}
\toprule
\multirow{2}{*}{Fold} & \multicolumn{2}{c|}{SEF} & \multicolumn{2}{c|}{PI3NN} & \multicolumn{2}{c}{PIVEN} \\
 & PICP & NMPIW & PICP & NMPIW & PICP & NMPIW \\
\midrule
1 & \textbf{0.960} & 0.294 & 0.860 & 0.405 & 0.960 & 0.361 \\
2 & \textbf{1.000} & 0.584 & 0.940 & 0.518 & 0.880 & 0.388 \\
3 & 0.900 & 0.302 & \textbf{0.940} & 0.471 & 0.900 & 0.420 \\
4 & 0.940 & \textbf{0.293} & \textbf{1.000} & 0.552 & 0.880 & 0.357 \\
5 & \textbf{1.000} & 0.677 & 0.960 & 0.571 & 0.940 & 0.491 \\
6 & 0.940 & 0.511 & 0.900 & 0.447 & \textbf{1.000} & 0.461 \\
7 & \textbf{0.940} & 0.471 & 0.820 & \textbf{0.393} & 0.820 & 0.443 \\
8 & \textbf{1.000} & 0.636 & 0.920 & 0.562 & 0.920 & 0.444 \\
9 & \textbf{0.940} & \textbf{0.330} & 0.940 & 0.518 & 0.920 & 0.428 \\
10 & \textbf{0.920} & 0.548 & 0.920 & 0.512 & 0.900 & 0.432 \\
\midrule
Mean & \textbf{0.954} & 0.465 & 0.920 & 0.495 & 0.912 & \textbf{0.422} \\
\bottomrule
\end{tabular}
\end{table}
Here, too, the SEF method shows the best average PICP, 0.954, compared to 0.92 for PI3NN and 0.912 for PIVEN. This is reinforced by the fact that SEF has the highest PICP value in 7 out of 10 folds. Again, the SEF method shows a better average PICP, 0.954 compared to 0.92 for PI3NN and 0.912 for PIVEN. This is reinforced by the fact that SEF has the highest PICP value in 7 out of 10 folds and that in 3 folds PICP was equal to 1.  The SEF method shows the second smallest average NMPIW value (0.465 vs. 0.422 of PIVEN) but with a significantly higher PICP value. From Friedman’s ranking method we conclude that there is no significant difference between the methods for both PICP and NMPIW, as p-values were equal to 0.074 and 0.302 respectively.

\section{Conclusions }\label{sec:conclusions}

In this work, we introduced the SEF (Shifting the Error Function) method, a novel approach to constructing prediction intervals (PIs) while addressing uncertainty quantification in neural network prediction.
Using both homoscedastic and heteroscedastic synthetic datasets, it was demonstrated that SEF could successfully produce narrow PIs without violating the predetermined confidence in variable noise.

The SEF method exhibits several notable strengths; among these is its ability to generate narrow PIs while consistently capturing the specified portion of data for a given confidence level, this capability was evident across the range of examples presented in Section~\ref{sec:Examples}. 
The main advantages of SEF method are:  a) it is universally applicable as it can be applied to any regression problem without any adjustments, b) it is simple to implement without complex or time-consuming procedures, c)  does not require complex network architectures, d) it has no computational cost since the only change in the implementation of a standard neural network is the addition of a constant $\mu$ to the loss function, e) it does not require the manual tuning of the optimal value of its hyperparameters as its only hyperparameter $\mu$ is automatically calculated by a very simple procedure by the first neural network that also gives the approximation value of the target value, g) compared to recent and recognized methods, it shows competitive results as in many cases it shows better values in the evaluation measures of PICP and NMPIW.

However, the method has certain weaknesses, such as: a) Sensitivity to extreme differences: The parameter $\mu$ and, consequently, the width of the PIs are influenced by extreme differences, especially when the noise $\varepsilon$ has heavy tails. This issue is common in other similar methods as well. b) Reapplication for different confidence levels: The method must be reapplied almost from the beginning for each different confidence level $\gamma$. c) Application in multi-output functions: The method becomes demanding when applied to functions with multiple outputs, as it needs to be applied separately for each output.

\section{Future Work}\label{sec:future}
Future research will focus on applying the SEF method to more complex datasets, exploring its applications to real-world problems, and evaluating its performance against other state-of-the-art methods. To this end, we plan to evaluate the SEF method on the UCI datasets, and compare its performance to other well established methods, including  PI3NN \cite{Liu_PI3NN_2022}, QD \cite{Pearce_2018}, PIVEN \cite{Simhayev_PIVEN_2020}, SQR \cite{Tagasovska_2019}, and DER \cite{Amini_2019}.
Additionally, it is worth investigating the relationship between the width of the generated prediction intervals (PIs) and the value of the parameter $\mu$, as well as the potential to apply the method with a smaller number of NNs. Moreover, a longer analysis of justification will be done by further analyzing some parts of the mathematical background of the method given briefly in this paper. 

\bibliographystyle{IEEEtran}
\bibliography{references}

\begin{thebibliography}{10}
\providecommand{\url}[1]{#1}
\csname url@samestyle\endcsname
\providecommand{\newblock}{\relax}
\providecommand{\bibinfo}[2]{#2}
\providecommand{\BIBentrySTDinterwordspacing}{\spaceskip=0pt\relax}
\providecommand{\BIBentryALTinterwordstretchfactor}{4}
\providecommand{\BIBentryALTinterwordspacing}{\spaceskip=\fontdimen2\font plus
\BIBentryALTinterwordstretchfactor\fontdimen3\font minus
  \fontdimen4\font\relax}
\providecommand{\BIBforeignlanguage}[2]{{%
\expandafter\ifx\csname l@#1\endcsname\relax
\typeout{** WARNING: IEEEtran.bst: No hyphenation pattern has been}%
\typeout{** loaded for the language `#1'. Using the pattern for}%
\typeout{** the default language instead.}%
\else
\language=\csname l@#1\endcsname
\fi
#2}}
\providecommand{\BIBdecl}{\relax}
\BIBdecl

\bibitem{Abdar_2021}
\BIBentryALTinterwordspacing
M.~Abdar, F.~Pourpanah, S.~Hussain, D.~Rezazadegan, L.~Liu, M.~Ghavamzadeh,
  P.~Fieguth, X.~Cao, A.~Khosravi, U.~R. Acharya, V.~Makarenkov, and
  S.~Nahavandi, ``A review of uncertainty quantification in deep learning:
  Techniques, applications and challenges,'' \emph{Information Fusion},
  vol.~76, p. 243–297, Dec. 2021. [Online]. Available:
  \url{http://dx.doi.org/10.1016/j.inffus.2021.05.008}
\BIBentrySTDinterwordspacing

\bibitem{Neal_1996}
\BIBentryALTinterwordspacing
R.~M. Neal, \emph{Bayesian Learning for Neural Networks}.\hskip 1em plus 0.5em
  minus 0.4em\relax Springer New York, 1996. [Online]. Available:
  \url{http://dx.doi.org/10.1007/978-1-4612-0745-0}
\BIBentrySTDinterwordspacing

\bibitem{Hwang_1997}
\BIBentryALTinterwordspacing
J.~T.~G. Hwang and A.~A. Ding, ``Prediction intervals for artificial neural
  networks,'' \emph{Journal of the American Statistical Association}, vol.~92,
  no. 438, pp. 748--757, Jun. 1997. [Online]. Available:
  \url{http://dx.doi.org/10.2307/2965723}
\BIBentrySTDinterwordspacing

\bibitem{de_Veaux_1998}
\BIBentryALTinterwordspacing
R.~D. de~Veaux, J.~Schumi, J.~Schweinsberg, and L.~H. Ungar, ``Prediction
  intervals for neural networks via nonlinear regression,''
  \emph{Technometrics}, vol.~40, no.~4, pp. 273--282, Nov. 1998. [Online].
  Available: \url{http://dx.doi.org/10.2307/1270528}
\BIBentrySTDinterwordspacing

\bibitem{Gawlikowski_2023}
\BIBentryALTinterwordspacing
J.~Gawlikowski, C.~R.~N. Tassi, M.~Ali, J.~Lee, M.~Humt, J.~Feng, A.~Kruspe,
  R.~Triebel, P.~Jung, R.~Roscher, M.~Shahzad, W.~Yang, R.~Bamler, and X.~X.
  Zhu, ``A survey of uncertainty in deep neural networks,'' \emph{Artificial
  Intelligence Review}, vol.~56, no.~S1, p. 1513–1589, Jul. 2023. [Online].
  Available: \url{http://dx.doi.org/10.1007/s10462-023-10562-9}
\BIBentrySTDinterwordspacing

\bibitem{Khosravi_2011a}
\BIBentryALTinterwordspacing
A.~Khosravi, S.~Nahavandi, D.~Creighton, and A.~F. Atiya, ``Comprehensive
  review of neural network-based prediction intervals and new advances,''
  \emph{{IEEE} Transactions on Neural Networks}, vol.~22, no.~9, p.
  1341–1356, Sep. 2011. [Online]. Available:
  \url{http://dx.doi.org/10.1109/tnn.2011.2162110}
\BIBentrySTDinterwordspacing

\bibitem{Jospin_2022}
\BIBentryALTinterwordspacing
L.~V. Jospin, H.~Laga, F.~Boussaid, W.~Buntine, and M.~Bennamoun, ``Hands-on
  bayesian neural networks—a tutorial for deep learning users,'' \emph{{IEEE}
  Computational Intelligence Magazine}, vol.~17, no.~2, p. 29–48, May 2022.
  [Online]. Available: \url{http://dx.doi.org/10.1109/mci.2022.3155327}
\BIBentrySTDinterwordspacing

\bibitem{Wang_2016}
\BIBentryALTinterwordspacing
H.~Wang and D.-Y. Yeung, ``Towards bayesian deep learning: A framework and some
  existing methods,'' \emph{IEEE Transactions on Knowledge and Data
  Engineering}, vol.~28, no.~12, p. 3395–3408, Dec. 2016. [Online].
  Available: \url{http://dx.doi.org/10.1109/tkde.2016.2606428}
\BIBentrySTDinterwordspacing

\bibitem{Gal_2016}
\BIBentryALTinterwordspacing
Y.~Gal and Z.~Ghahramani, ``Dropout as a bayesian approximation: Representing
  model uncertainty in deep learning,'' in \emph{Proceedings of The 33rd
  International Conference on Machine Learning}, ser. Proceedings of Machine
  Learning Research, M.~F. Balcan and K.~Q. Weinberger, Eds., vol.~48.\hskip
  1em plus 0.5em minus 0.4em\relax New York, New York, USA: PMLR, 20--22 Jun
  2016, pp. 1050--1059. [Online]. Available:
  \url{https://proceedings.mlr.press/v48/gal16.html}
\BIBentrySTDinterwordspacing

\bibitem{Pearce_2020}
\BIBentryALTinterwordspacing
T.~Pearce, F.~Leibfried, and A.~Brintrup, ``Uncertainty in neural networks:
  Approximately bayesian ensembling,'' in \emph{Proceedings of the Twenty Third
  International Conference on Artificial Intelligence and Statistics}, ser.
  Proceedings of Machine Learning Research, S.~Chiappa and R.~Calandra, Eds.,
  vol. 108.\hskip 1em plus 0.5em minus 0.4em\relax PMLR, 26--28 Aug 2020, pp.
  234--244. [Online]. Available:
  \url{https://proceedings.mlr.press/v108/pearce20a.html}
\BIBentrySTDinterwordspacing

\bibitem{Hu_2019}
\BIBentryALTinterwordspacing
R.~Hu, Q.~Huang, S.~Chang, H.~Wang, and J.~He, ``The {MBPEP}: a deep ensemble
  pruning algorithm providing high quality uncertainty prediction,''
  \emph{Applied Intelligence}, vol.~49, no.~8, p. 2942–2955, Feb. 2019.
  [Online]. Available: \url{http://dx.doi.org/10.1007/s10489-019-01421-8}
\BIBentrySTDinterwordspacing

\bibitem{Fersini_2014}
\BIBentryALTinterwordspacing
E.~Fersini, E.~Messina, and F.~Pozzi, ``Sentiment analysis: Bayesian ensemble
  learning,'' \emph{Decision Support Systems}, vol.~68, p. 26–38, Dec. 2014.
  [Online]. Available: \url{http://dx.doi.org/10.1016/j.dss.2014.10.004}
\BIBentrySTDinterwordspacing

\bibitem{Tagasovska_2019}
\BIBentryALTinterwordspacing
N.~Tagasovska and D.~Lopez-Paz, ``Single-model uncertainties for deep
  learning,'' in \emph{Proceedings of the 33rd International Conference on
  Neural Information Processing Systems}, H.~Wallach, H.~Larochelle,
  A.~Beygelzimer, F.~d\textquotesingle Alch\'{e}-Buc, E.~Fox, and R.~Garnett,
  Eds., vol.~32.\hskip 1em plus 0.5em minus 0.4em\relax Red Hook, NY, USA:
  Curran Associates Inc., 2019. [Online]. Available:
  \url{https://proceedings.neurips.cc/paper_files/paper/2019}
\BIBentrySTDinterwordspacing

\bibitem{Khosravi_2011b}
\BIBentryALTinterwordspacing
A.~Khosravi, S.~Nahavandi, D.~Creighton, and A.~F. Atiya, ``Lower upper bound
  estimation method for construction of neural network-based prediction
  intervals,'' \emph{IEEE Transactions on Neural Networks}, vol.~22, no.~3, p.
  337–346, Mar. 2011. [Online]. Available:
  \url{http://dx.doi.org/10.1109/tnn.2010.2096824}
\BIBentrySTDinterwordspacing

\bibitem{Pearce_2018}
\BIBentryALTinterwordspacing
T.~Pearce, A.~Brintrup, M.~Zaki, and A.~Neely, ``High-quality prediction
  intervals for deep learning: A distribution-free, ensembled approach,'' in
  \emph{Proceedings of the 35th International Conference on Machine Learning},
  ser. Proceedings of Machine Learning Research, J.~Dy and A.~Krause, Eds.,
  vol.~80.\hskip 1em plus 0.5em minus 0.4em\relax PMLR, 10--15 Jul 2018, pp.
  4075--4084. [Online]. Available:
  \url{https://proceedings.mlr.press/v80/pearce18a.html}
\BIBentrySTDinterwordspacing

\bibitem{Simhayev_2022}
\BIBentryALTinterwordspacing
E.~Simhayev, G.~Katz, and L.~Rokach, ``Integrated prediction intervals and
  specific value predictions for regression problems using neural networks,''
  \emph{Knowledge-Based Systems}, vol. 247, p. 108685, Jul. 2022. [Online].
  Available: \url{http://dx.doi.org/10.1016/j.knosys.2022.108685}
\BIBentrySTDinterwordspacing

\bibitem{Simhayev_PIVEN_2020}
\BIBentryALTinterwordspacing
------, ``{PIVEN}: A deep neural network for prediction intervals with specific
  value prediction,'' 2020. [Online]. Available:
  \url{https://arxiv.org/abs/2006.05139}
\BIBentrySTDinterwordspacing

\bibitem{Liu_PI3NN_2022}
\BIBentryALTinterwordspacing
S.~Liu, P.~Zhang, D.~Lu, and G.~Zhang, ``{PI3NN}: Out-of-distribution-aware
  prediction intervals from three neural networks,'' 2021. [Online]. Available:
  \url{https://doi.org/10.48550/arXiv.2108.02327}
\BIBentrySTDinterwordspacing

\bibitem{MathWorks}
\BIBentryALTinterwordspacing
{MathWorks}, \emph{Outlier Detection Using Quantile Regression}, MathWorks,
  2024. [Online]. Available:
  \url{https://www.mathworks.com/help/stats/outlier-detection-using-quantile-regression.html}
\BIBentrySTDinterwordspacing

\bibitem{Amini_2019}
\BIBentryALTinterwordspacing
A.~Amini, W.~Schwarting, A.~Soleimany, and D.~Rus, ``Deep evidential
  regression,'' 2019. [Online]. Available:
  \url{https://doi.org/10.48550/arXiv.1910.02600}
\BIBentrySTDinterwordspacing

\end{thebibliography}

\end{document}